\documentclass[10pt,twocolumn,letterpaper]{article}

\usepackage{cvpr}
\usepackage{times}
\usepackage{epsfig}
\usepackage{graphicx}
\usepackage{amsmath}
\usepackage{amssymb}
\usepackage[table]{xcolor}
\usepackage{booktabs}

\usepackage[breaklinks=true,bookmarks=false]{hyperref}

\cvprfinalcopy 

\def\cvprPaperID{****} 

\begin{document}

\title{Tree-structured Kronecker Convolutional Network for Semantic Segmentation}
\author{Tianyi Wu$^{1,2}$, Sheng Tang$^{1,}$\thanks{Corresponding author: Sheng Tang (ts@ict.ac.cn)}\, , Rui Zhang$^{1,2}$, Juan Cao$^{1}$, Jintao Li$^{1}$\\
$^1$Institute of Computing Technology, Chinese Academy of Sciences, Beijing, China. \\
$^2$ University of Chinese Academy of Sciences, Beijing, China. \\
{\tt\small {\{wutianyi, ts, zhangrui, caojuan, jtli\}}@ict.ac.cn}
}

\maketitle

\begin{abstract}
Most existing semantic segmentation methods employ atrous convolution to enlarge the receptive field of filters, but neglect partial information. To tackle this issue, we firstly propose a novel Kronecker convolution which adopts Kronecker product to expand the standard convolutional kernel for taking into account the partial feature neglected by atrous convolutions. Therefore, it can capture partial information and enlarge the receptive field of filters simultaneously without introducing extra parameters. Secondly, we propose Tree-structured Feature Aggregation (TFA) module which follows a recursive rule to expand and forms a hierarchical structure. Thus, it can naturally learn representations of multi-scale objects and encode hierarchical contextual information in complex scenes. Finally, we design Tree-structured Kronecker Convolutional Network (TKCN) which employs Kronecker convolution and TFA module. Extensive experiments on three datasets, PASCAL VOC 2012, PASCAL-Context and Cityscapes, verify the effectiveness of our proposed approach. We make the code and the trained model publicly available at \textcolor[rgb]{1,0,0}{https://github.com/wutianyiRosun/TKCN}.

\end{abstract}

\begin{figure}[t]
\centering
\vspace{-0.2cm}
\includegraphics[width=0.9\linewidth]{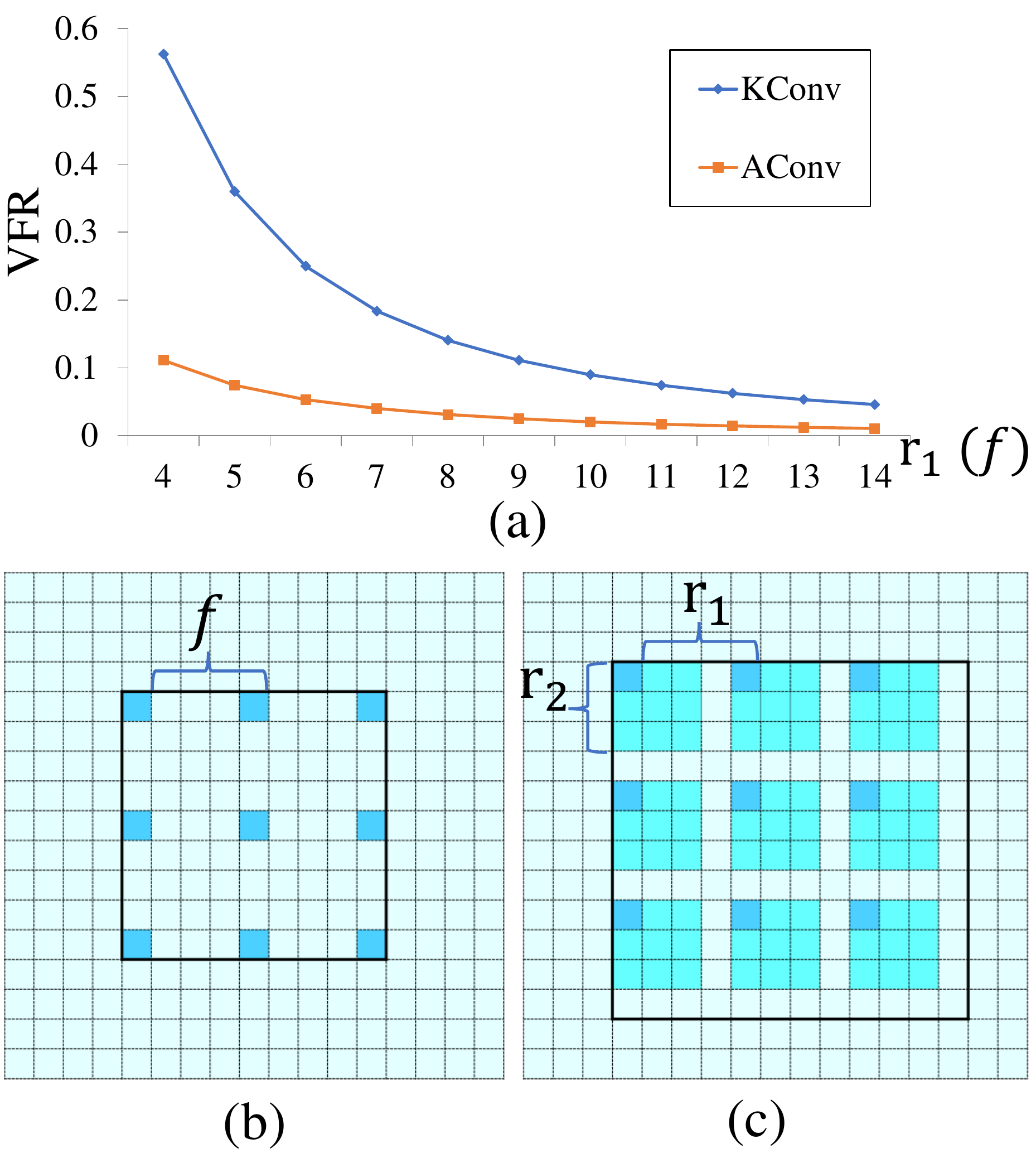}
\caption{(a) Curves of the VFR for atrous convolutions (AConv) and Kronecker convolutions (KConv) with different rates. Inter-dilating factor $r_1$ of Kronecker convolution is equivalent to rate $f$ of atrous convolution. Note that $r_2 = 3$. (b) Atrous convolution with rate $f=4$. (c) Kronecker convolution with inter-dilating factor $r_1=4$, intra-sharing factor $r_2=3$. In (b) and (c), cells in the black boxes represent feature vectors in convolutional patches, while cyan and blue cells represent feature vectors involved in computation.}
\vspace{-15pt}
\label{fig:fig1}
\end{figure}

\section{Introduction}
Semantic segmentation is a significant challenge for computer vision. The goal of semantic segmentation is to assign one of the semantic labels to each pixel in an image. Current segmentation models \cite{chen2016deeplab,shelhamer2017fully} based on Deep Convolutional Neural Networks (DCNNs) achieve good performances on several semantic segmentation benchmarks \cite{everingham2015pascal,cordts2016cityscapes}, such as Fully Convolutional Networks (FCNs) \cite{shelhamer2017fully}. These models transfer classification networks \cite{simonyan2014very,he2016deep} pre-trained on ImageNet dataset \cite{russakovsky2015imagenet} to generate segmentation predictions through removing max-pooling, altering fully connected layers and adding deconvolutional layers. More recently, employing atrous convolutions, also named dilated convolutions \cite{yu2015multi}, instead of standard convolutions in some layers of FCNs has become the mainstream, since atrous convolutions can enlarge the field of view and maintain the resolution of feature maps. Although atrous convolutions show good performances in semantic segmentation, it lacks the capability of capturing partial information. To illustrate this issue better, we define Valid Feature Ratio (VFR) as the ratio of the number of feature vectors involved in the computation to that of all feature vectors in the convolution patch. VFR can measure the utilization ratio of features in convolutional patches. As shown in Fig.~\ref{fig:fig1} (a), the VFR of atrous convolutions is relatively low, which means much more partial information is neglected. As shown in Fig.~\ref{fig:fig1} (b), we can observe that atrous convolutions lose important partial information when employing large rate $f$. Typically, only 9 out of 81 feature vectors in the convolutional patch are involved in the computation.
Specially, when the rate is extremely large and exceeds the sizes of feature maps, the 3$\times$3 filters will degenerate to 1$\times$1 filters without capturing the global contextual information, since only the center filter branch is effective.

In order to address the above problems, we propose Kronecker convolutions inspired by Kronecker product in computational and applied mathematics \cite{henderson1983history}. Our proposed Kronecker convolutions can not only inherit the advantages of atrous convolutions, but also mitigate its limitation. The proposed Kronecker convolutions employ Kronecker product to expand standard convolutional kernel
 so that feature vectors neglected by atrous convolutions can be captured, as shown in Fig.~\ref{fig:fig1} (c). There are two factors in Kronecker convolution, inter-dilating factor $r_1$ and intra-sharing factor $r_2$. On one hand, the inter-dilating factor controls the number of holes inserted into kernels. Therefore, Kronecker convolutions have the capability of enlarging the field of view and maintain the resolution of feature maps, namely Kronecker convolutions can inherit the advantages of atrous convolutions. On the other hand, the intra-sharing factor controls the size of subregions to capture feature vectors and share filter vectors. Thus, Kronecker convolutions can consider partial information and increase VFR without increasing the number of parameters, as shown in Fig.~\ref{fig:fig1} (a) and (c).

Furthermore, scenes in images have hierarchical structures, which can be decomposed into small-range scenes or local scenes (such as a single object), middle-range scenes (such as multiple objects) and large-range scenes (such as the whole image). How to efficiently capture hierarchical contextual information in complex scenes is significant to semantic segmentation and remains a challenge. Based on this observation, we propose Tree-structured Feature Aggregation (TFA) module to encode hierarchical context information, which is beneficial to better understand complex scenes and improve segmentation accuracy.
Our TFA module follows a recursive rule to expand, and forms a tree-shaped and hierarchical structure.
Each layer in TFA model has two branches, one branch preserves features of the current region, and the other branch aggregates spatial dependencies within a larger range.
The proposed TFA module has two main advantages:
(1) Oriented for the hierarchical structures in the complex scenes, TFA module can capture the hierarchical contextual information effectively and efficiently;
(2) Compared with the existing multi-scale feature fusion methods based on preset scales and relied on inherent network structures, TFA module can naturally learn representations of multi-scale objects by the tree-shaped structure.

Based on the above observation, we propose Tree-structured Kronecker Convolutional Network (TKCN) for semantic segmentation, which employs Kronecker convolution and TFA module
to form a unified framework. We perform experiments on three popular semantic segmentation benchmarks, including PASCAL VOC 2012, Cityscapes and PASCAL-Context.
Experimental results verify the effectiveness of our proposed approaches. Our main contributions can be summarized into three aspects:
\begin{itemize}
\item We propose Kronecker convolutions, which can effectively capture partial detail information and enlarge the field of view simultaneously, without introducing extra parameters.
\item We develop Tree-structured Feature Aggregation module to capture hierarchical contextual information and represent multi-scale objects, which is beneficial for better understanding complex scenes.
\item Without any post-processing steps, our designed TKCN achieves impressive results on the benchmarks of PASCAL VOC 2012, Cityscapes and PASCAL-Context.
\end{itemize}

\section{Related Work}
In this section, we firstly overview the using of Kronecker product in deep learning and popular semantic segmentation approaches, and then introduce related approaches of two aspects of semantic segmentation, including Conditional Random Fields (CRFs) and Multi-scale Feature Fusion.

\noindent\textbf{Kronecker Product}\quad
KFC \cite{zhou2015exploiting} uses Kronecker product to exploit the local structures within convolution and fully-connected layers, by replacing the large weight matrices and by combinations of multiple Kronecker products of smaller matrices, which can approximate the weight matrices of the fully connected layer. In contrast to them, we employ Kronecker product to expand the standard convolutional kernel for enlarging the receptive field of filters, and capturing partial information neglected by atrous convolutions.

\begin{figure*}[tp]
\centering
    \includegraphics[width=0.85\linewidth]{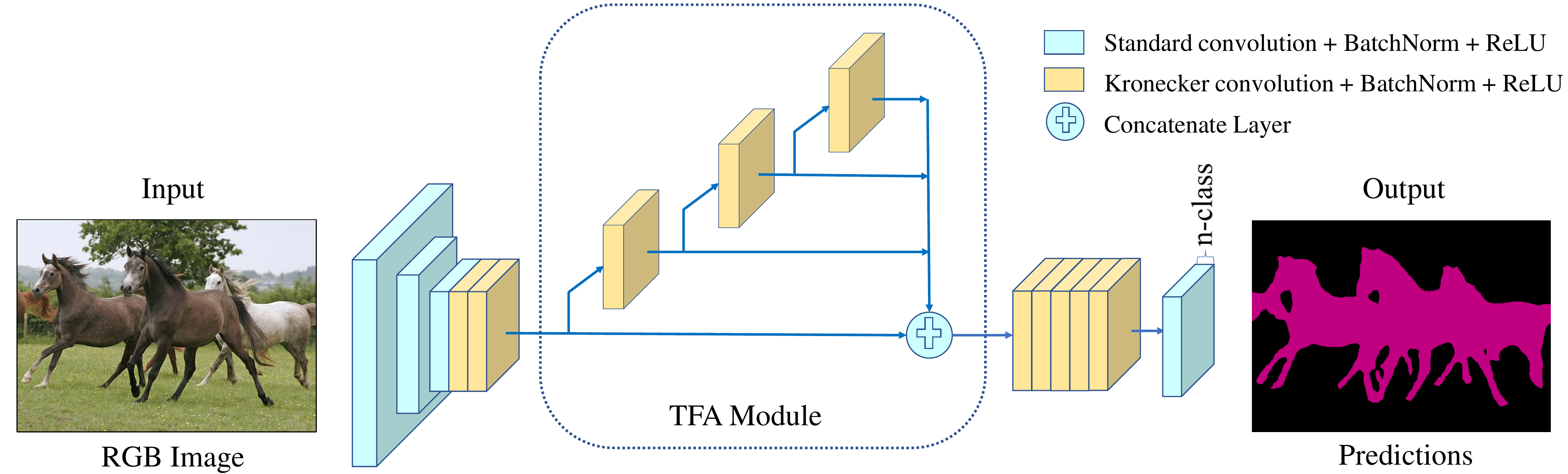}
    \caption{Architecture of the proposed TKCN. We employ Kronecker convolutions in ResNet-101 `Res4' and `Res5'. Tree-structured Feature Aggregation module is implemented after the last layer of `Res5'.}
        \label{fig:fig2}
\end{figure*}

\noindent\textbf{Semantic Segmentation}\quad
Semantic segmentation is a fundamental task in computer vision.
Recently, approaches based on Deep Convolutional Neural Networks \cite{krizhevsky2012imagenet,simonyan2014very,he2016deep} achieve remarkable progress in semantic segmentation task, such as DeconvNets \cite{noh2015learning}, DeepLab \cite{chen2016deeplab} and FCNs \cite{shelhamer2017fully}. FCNs transfer the networks of image classification for pixel-level labeling. DeconvNets employ multiple deconvolution layers to enlarge feature maps and generate whole-image predictions. DeepLab methods use atrous convolutions to enlarge the receptive fields so as to capture contextual information. Following these structures, many frameworks are proposed to further improve the accuracy of semantic segmentation.

\noindent\textbf{Conditional Random Fields}\quad
One common approach to capture fine-grained details and refine the segmentation predictions is CRFs, which are suitable for capturing long-range dependencies and fine local details. CRFasRNN \cite{zheng2015conditional} reformulates DenseCRF with pairwise potential functions and unrolls the mean-field steps as recurrent neural networks, which composes a uniform framework and can be learned end-to-end. Differently, DeepLab frameworks \cite{chen2016deeplab} use DenseCRF \cite{krahenbuhl2011efficient} as post-processing. After that, many approaches combine CRFs and DCNNs in the uniform frameworks, such as combining Gaussian CRFs \cite{vemulapalli2016gaussian} and specific pairwise potentials \cite{jampani2016learning}. In contrast, some other approaches directly learn pairwise relationships.  SPN \cite{liu2017learning} constructs a row/column linear propagation model to capture dense, global pairwise relationships in an image, and Spatial CNN \cite{pan2017spatial} learns the spatial relationships of pixels across rows and columns in an image.
While these approaches achieve remarkable improvement, they increase the overall computational complexity of the networks.

\noindent\textbf{Multi-scale Feature Fusion}\quad
Since objects in scene images have various sizes, multi-scale feature fusion is widely used in semantic segmentation approaches for learning features of multiple scales. Some approaches aggregate features of multiple meddle layers.
The original FCNs \cite{shelhamer2017fully} utilize skip connections to perform late fusion. Hypercolumn \cite{hariharan2015hypercolumns} merges features from middle layers to learn dense classification layers.
RefineNet \cite{lin2017refinenet} proposes to pool features with multiple window sizes and fuses them together with residual connections and learnable weights.
Some methods obtain multi-scale features from inputs, such as utilizing a Laplacian pyramid \cite{farabet2013learning}, employing multi-scale inputs sequentially from coarse-to-fine \cite{pinheiro2014recurrent}, or simply resizing input images into multiple sizes \cite{lin2017exploring}.
Some other approaches propose feature pyramid modules. DeepLab-v2 \cite{chen2016deeplab} employs four parallel atrous convolutional layers of different rates to capture objects and context information of multiple scales. PSPNet \cite{Zhao_2017_CVPR} performs spatial pooling at four grid scales. More recently, DFN \cite{Yu_2018_CVPR} propose a Smooth Network for fusing feature maps across different stages, and CCL \cite{Ding_2018_CVPR} propose a scheme of gated sum to selectively aggregate multi-scale features for each spatial position. Most multi-scale feature fusion methods are compromised by preset scales or relying on inherent network structure.

In this paper, we propose Kronecker convolutions to capture partial information neglected by atrous convolutions. Different from the computationally expensive CRF-based approaches and feature fusion methods with manually preset scales, we propose the TFA module to efficiently aggregate features of multiple scales through a tree-shaped structure and the recursive rule.

\section{Proposed Approaches}
\label{sec:blind}

In this paper, we design the Tree-structured Kronecker Convolutional Network (TKCN) for semantic segmentation. As illustrated in Fig.~\ref{fig:fig2}, TKCN employs our proposed Kronecker convolution and TFA module.
In the following, firstly we formulate the proposed Kronecker convolutions and explain why they can capture partial detail information when enlarging the receptive fields. Then we present the TFA module and analyze how it can efficiently aggregate hierarchical contextual information.

\subsection{Kronecker Convolution}
Inspired by Kronecker product in computational and applied mathematics, we explore a novel Kronecker convolution whose kernels are transformed by performing Kronecker product.

First of all, we provide a brief review of Kronecker product. If $ \mathbf{A}$ is a $m \times n$ matrix and $\mathbf{B}$ is a $ r \times s$ matrix, the Kronecker product $ \mathbf{A} \otimes \mathbf{B} $ is the $mr \times ns$ matrix:
\begin{align}
\mathbf{A} \otimes \mathbf{B}= \begin{bmatrix} a_{11} \mathbf{B} & \cdots & a_{1n}\mathbf{B} \\ \vdots & \ddots & \vdots \\ a_{m1} \mathbf{B} & \cdots & a_{mn} \mathbf{B} \end{bmatrix}.
\label{formulation0}
\end{align}


For a standard convolution, it takes the input feature maps $A\in\mathbb{R}^{H_A\times W_A\times C_A}$ and outputs feature maps $B\in\mathbb{R}^{H_B\times W_B\times C_B}$, where $H_A$, $W_A$, $C_A$, $H_B$, $W_B$, $C_B$ are the widths, heights and channels of $A$ and $B$ respectively. The kernel of the standard convolution is $K\in\mathbb{R}^{C_B\times C_A\times (2k+1)\times (2k+1)}$ and the bias is $b\in\mathbb{R}^{C_B}$. Any feature vector $B^t\in\mathbb{R}^{C_B}$ in $B$ at position $t$ is the multiplication of kernel $K$ and the associated convolutional patch $X^t$ in $A$, where $t\in[1,W_B\times H_B]\cap\mathbb{Z}$. $X^t$ is a $(2k+1)\times (2k+1)$ square with the center of $(p^t,q^t)$, where $p^t\in[1,H_A]\cap\mathbb{Z}$ and $q^t\in[1,W_A]\cap\mathbb{Z}$ are coordinates in $A$. So the coordinates of feature vectors in $X^t$ are:
\begin{align}
x_{ij} = p^t+i, \quad y_{ij}=q^t+j,
\end{align}
where $i,j\in[-k,k]\cap\mathbb{Z}$. Let $X^t_{ij}=X^t(x_{ij},y_{ij})\in\mathbb{R}^{C_A}$, $K_{ij}=K(i,j)\in\mathbb{R}^{C_B\times C_A}$, the convolutional operator can be formulated as matrix multiplication:
\begin{align}
B^t=\sum_{i,j}K_{ij}X^t_{ij}+b.
\end{align}

For our proposed Kronecker convolution, we introduce a transformation matrix $F$ and enlarge the kernel $K$ through computing Kronecker product of $ F$ and $K$. We set $F$ as a fixed $r_1\times r_1$ matrix. Inter-dilating factor $r_1$ can control the dilation rate of the convolutions. Therefore, the kernel of Kronecker convolution will expand from $K$ of $(2k+1)\times (2k+1)$  to $K'$ of $(2k+1)r_1\times (2k+1)r_1$.
In order to avoid bringing extra parameters in the Kronecker convolution, we simply set $F$ as the combination of a matrix $\mathbf{I}$ and zero matrix $\mathbf{O}$, where $\mathbf{I}$ is a $r_2 \times r_2$ square matrix which has all the element values of 1. We denote the intra-sharing factor as $r_2$ ($1\leq r_2\leq r_1$), which controls the size of subregions to capture feature vectors and share filter vectors. Thus, the kernel $K'$ of Kronecker convolution can be formulated as:

\begin{align}
\begin{split}
K'(c_2,c_1) = K(c_2,c_1)\otimes F, \\
 \quad F=\begin{bmatrix}
\mathbf{ I_{r_2\times r_2}}, & \\
 & \mathbf{ O_{(r_1-r_2) \times  (r_1-r_2)}}
\end{bmatrix},
\end{split}
\end{align}
where $c_2\in[1,C_B]\cap\mathbb{Z}$, $c_1\in[1,C_A]\cap\mathbb{Z}$. Correspondingly, the associated convolutional patch in $A$, denoted as $Y^t$ , will also expand to a square of $(2k+1)r_1\times (2k+1)r_1$. Coordinates of feature vectors involved in computation in $Y^t$ are:
\begin{align}
x_{ijuv} = p^t+i r_1+u, \quad y_{ijuv} = q^t+j r_1+v,
\label{eq1}
\end{align}
where $i,j\in[-k,k]\cap\mathbb{Z}$, $u,v\in[0,r_2-1]\cap\mathbb{Z}$. Let $Y^t_{ijuv}=Y^t(x_{ijuv},y_{ijuv})\in\mathbb{R}^{C_A}$,  $K'_{ijuv}=K'(ir_1 +u,jr_1 +v)\in\mathbb{R}^{C_A}$. Therefore, the operator of the Kronecker convolution can be formulated as:
\begin{align}
B^t=\sum_{i,j,u,v}K'_{ijuv}Y^t_{ijuv}+b
   =\sum_{i,j}K_{ij}\sum_{u,v}Y^t_{ijuv}+b.
\label{eq2}
\end{align}

\begin{figure}[t]
\centering
\includegraphics[width=0.8\linewidth]{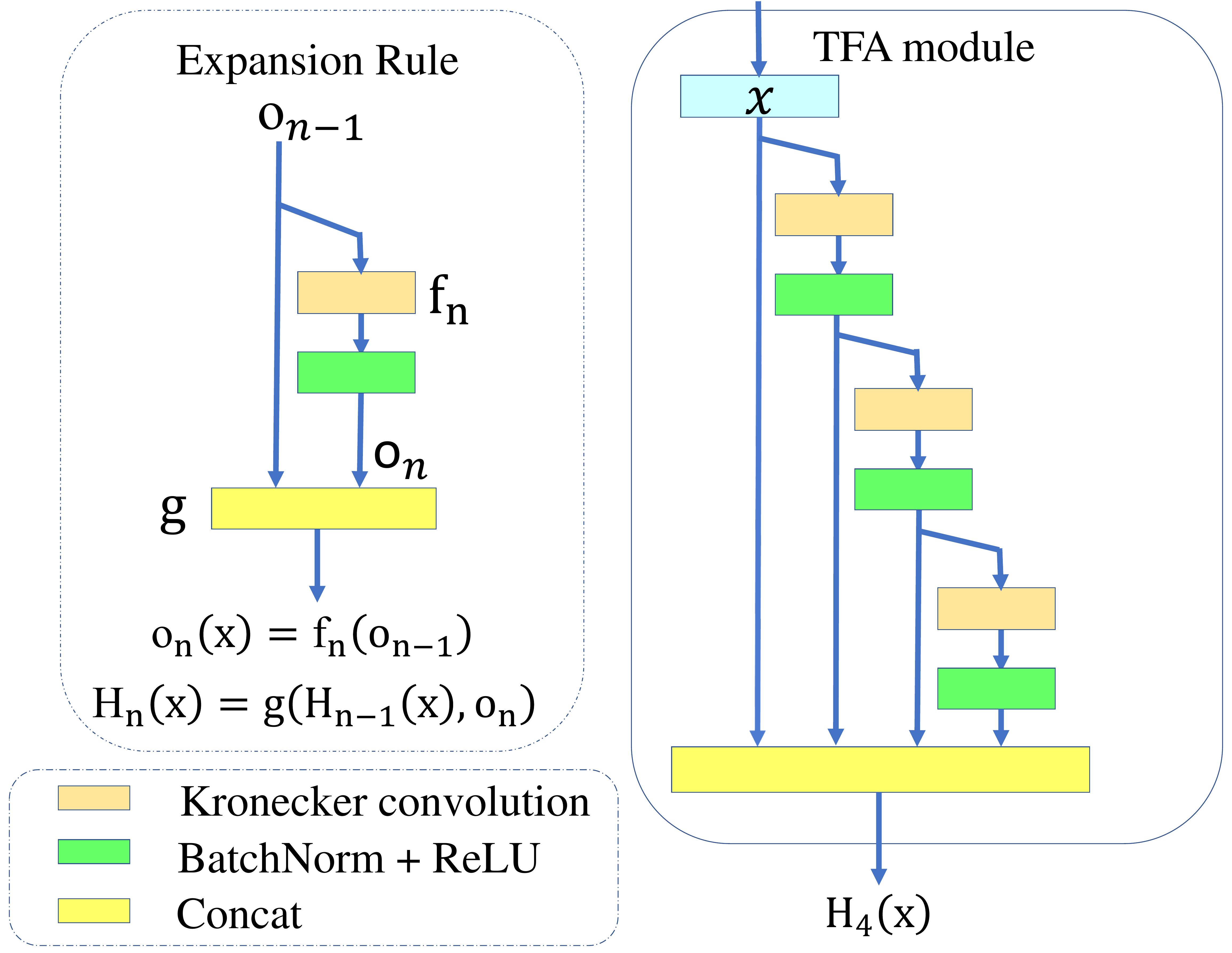}
\caption{Left: A simple expansion rule generates a TFA architecture. Right: Tree-structured Feature Aggregation module. }
\vspace{-10pt}
\label{fig:fig4}
\end{figure}

Compared with atrous convolutions which simply insert zeros to expand kernels, Kronecker convolutions expand kernels through Kronecker product with transformation matrix $ F$. The inter-dilating factor $r_1$ controls the dilation rate of kernels. According to Eqn.~\eqref{eq1}, when $r_1$ becomes larger, the convolutional patches zoom in so that the receptive fields will be enlarged correspondingly.
Since $ F$ only contains values of ones and zeros, no more parameters are introduced in Kronecker convolutions. Moreover, because $ F$ has a submatrix of $r_2\times r_2$ identity matrix, Kronecker convolutions can capture local contextual information ignored by atrous convolutions. As shown in  Eqn.~\eqref{eq2}, each kernel branch $K_{ij}$ has the capability of aggregating features in a $r_2\times r_2$ subregion. The VFR of Kronecker convolutions is $r_2^2/r_1^2$, while atrous convolutions with the same rate $r_1$ have the VFR of $1/r_1^2$. It is clear that VFR of Kronecker convolutions is larger than atrous convolutions since $r_2\geq 1$. When $r_1=r_2$, the VFR of Kronecker convolutions will be 100\%. In conclusion, our proposed Kronecker convolutions can capture partial information and enlarge the field of view simultaneously without increasing extra parameters.

The proposed Kronecker convolutions can be treated as the generalization of standard convolutions and atrous convoltuions. If $r_2=1$, Kronecker convoltions will degenerate to atrous convolutions, since the kernel will change to:
\begin{align}
K''(c_2,c_1) = K(c_2,c_1)\otimes  F, \quad  F=\begin{bmatrix}
1 & 0 & \cdots & 0\\
0 & 0 & \cdots & 0\\
\vdots & \vdots & \ddots & \vdots \\
0 & 0 & \cdots & 0
\end{bmatrix}.
\end{align}
Therefore, the formulation of Eqn.~\eqref{eq1} and \eqref{eq2} will change to:
\begin{align}
x_{ij} = p^t+i r_1, \quad y_{ij} = q^t+j r_1.
\end{align}
Let $K''_{ij}=K''(ir_1 ,jr_1)\in\mathbb{R}^{C_A}$, corresponding convolutions patch in $A$ is $Z^t_{ij}=Z^t(x_{ij},y_{ij})\in\mathbb{R}^{C_A}$. So
\begin{align}
B^t=\sum_{i,j}K_{ij}Z^t_{ij}+b.
\end{align}
 Additionally, if $r_1=r_2=1$, Kronecker convolutions will degenerate to standard convolutions.


\subsection{Tree-structured Feature Aggregation Module}

In order to capture hierarchical context information and represent objects of multiple scales in complex scenes, we propose TFA module. TFA module takes the features extracted by the backbone network as the input. TFA module follows an expansion and stacking rule to efficiently encode multi-scale features. As illustrated in the left subfigure of Fig.~\ref{fig:fig4}, in each expansion step, the input is duplicated to two branches. One branch preserves features of the current scale, and the other branch explores spatial dependencies within a larger range. Simultaneously, output features of the current step are stacked with previous features through concatenation. This expansion and stacking rule can be formulated as:
\begin{align}
o_n = f_n(o_{n-1}), \quad H_n(x) = g(H_{n-1}(x), o_n),
\end{align}
where $o_0=x$, $o_n$ is the output of step n, $f_n$ denotes operators implemented in step n, $H_n$ is the result of TFA module with n steps, and $g$ represents the concatenation operator. In the proposed TFA module, we employ Kronecker convolutions with different inter-dilating factors and intra-sharing factors to capture multi-scale features, followed by Batch Normalization and ReLU layers. Finally, the features of all the branches will be aggregated. As shown in the right subfigure of Fig.~\ref{fig:fig4}, in our experiments, we exploit TFA module with three expansion steps, so the features from all branches are concatenated finally. Particularly, to make a trade-off between computational complexity and model capability, we reduce the output channel of each convolutional layers in TFA module as $C/4$ if the input feature maps of TFA module has the channel of $C$.

Following the above expansion and stacking rule, TFA module forms a tree-shaped and hierarchical structure, which can effectively and efficiently capture hierarchical contextual information and aggregate features from multiple scales. Moreover, features learned from the previous steps can be re-explored in the subsequent steps, which is superior to the existing parallel structure with multiple individual branches.

\section{Experiments}


In this section, we perform comprehensive experiments on three semantic segmentation benchmarks to show the effectiveness of our proposed approaches, including PASCAL VOC 2012 \cite{everingham2015pascal}, Cityscapes \cite{cordts2016cityscapes} and PASCAL-Context \cite{mottaghi2014role}.

\begin{table}[t]
\small
\centering

\caption{ Evaluation results of Kronecker convolution (KConv) with different intra-sharing factor $r_2$ on PASCAL VOC 2012 validation set. }
\label{table:headings12}
\begin{tabular}{p{1.2cm}<{\centering}p{1.2cm}<{\centering}p{1.2cm}<{\centering}p{1.2cm}<{\centering}}

\toprule[1.5pt]

$r_1$  & $r_2$   & \bf{mIoU(\%)} & \bf{Acc(\%)}\\
\midrule
 6 &  1   &77.03  & 94.97 \\
 6 &  3   &78.37  & 95.25 \\
  6 &  5  &78.75  & 95.36 \\
\midrule
 10 &  1   &78.01  & 95.17 \\
 10  &  3  &78.53  & 95.24 \\
 10  &  5   &78.93  & 95.34 \\
 10  & 7   &79.50  & 95.53 \\
 10  &  9   &79.71  & 95.54 \\
\bottomrule[1.5pt]
\end{tabular}
\end{table}

\subsection{Experimental Settings}

\noindent\textbf{PASCAL VOC 2012 Dataset}\quad
The PASCAL VOC 2012 segmentation benchmark \cite{everingham2015pascal} contains 20 foreground object categories and 1 background class. The original dataset involves 1, 464 training images, 1, 449 validation images, and 1, 456 test images. Extra annotations from \cite{hariharan2011semantic} are provided to augment the training set to 10, 582 images. The performance is measured by pixel intersection-over-union (IoU) averaged across the 21 classes.

\noindent\textbf{Cityscapes Dataset}\quad
The Cityscapes datasets \cite{cordts2016cityscapes} contains 5, 000 images collected in street scenes from 50 different cities. The dataset is divided into three subsets, including 2, 975 images in training set, 500 images in validation set and 1, 525 images in test set. High-quality pixel-level annotations of 19 semantic classes are provided in this dataset. Intersection over Union (IoU) averaged over all the categories is adopted for evaluation.

\noindent\textbf{PASCAL-Context Dataset}
The PASCAL-Context dataset \cite{mottaghi2014role} involves 4, 998 images in training set and 5, 105 images in validation set. It provides detailed semantic labels for the whole scene. Our proposed models are evaluated on the most frequent 59 categories and 1 background class.

\subsubsection{Implementation Details}

We take ResNet-101\cite{he2016deep} as our baseline model, which employ atrous convolutions with $f=2$ and $f=4$ in layers of `Res4' and `Res5', respectively. So the resolution of the predictions can be enlarged from $ 1/32$ to $1/8$. Our loss function is the sum of cross-entropy terms for each spatial position in the output score map, ignoring the unlabeled pixels. All the experiments are performed on the Caffe platform. We employ the ``poly'' learning rate policy, in which we set the base learning rate to $0.001$ and power to $0.9$. Momentum and weight decay are set to $0.9$ and $0.0001$ respectively. For data augmentation, we employ random mirror and random resize between 0.5 and 2 for all training samples.

\subsection{Ablation Studies}
We evaluate the effectiveness of the two proposed components, Kronecker convolution and TFA module. All the experiments of ablation studies are conducted on PASCAL VOC 2012 dataset.

\begin{table}[t]
\small
\centering

\caption{ Comparison between Kronecker convolutions (KConv) and atrous convolutions (AConv) on PASCAL VOC 2012 validation set.}
\label{table:headings1}
\begin{tabular}{lcccc}
\toprule[1.5pt]

 Method & $r_1$  & $r_2$ & \bf{ mIoU (\%)} & \bf{Acc (\%)}\\
\midrule

AConv (Baseline)  &  4  & 1  &75.98  & 94.80\\
KConv              &  4 & 3   &76.70  & 94.98\\
\hline
AConv               &  6  &  1  &77.03  & 94.97 \\
KConv               &  6  &  5   &78.75  & 95.36 \\
\hline

AConv               &  8  &  1   &78.14  & 95.19 \\
KConv              &  8  &  5   &78.81  & 95.30 \\

\hline
AConv               &  10 &  1   &78.01  & 95.17 \\
KConv               &  10 &  7   &79.50  & 95.53 \\

\hline
AConv               &  12 &  1   &78.18  & 95.21\\
KConv              &  12 &  9   &79.79  & 95.53\\
\bottomrule[1.5pt]
\end{tabular}
\end{table}

\subsubsection{Ablation Study for Kronecker Convolution}
In order to analyze the effectiveness of Kronecker convolutions, we employ Kronecker convolutions with different ${r_1}$  and ${r_2}$ factors in ResNet-101 `Res5'. Firstly, we analyze the effect of varying intra-sharing factor $r_2$. As shown in Tab.~\ref{table:headings12}, we fix the inter-dilating factor with $r_1=6$ and change $r_2$ from 1 to 5, the mean IoU is continuously improved from 77.03\% to 78.75\%. Similar results are gained with a fixed $r_1=10$, in which the mean IoU increases from 78.01\% to 79.71\% as $r_2$ increases from 1 to 9. These results show that, with the increase of $r_2$, more partial information in convolutional patches can be captured, so that the mean IoU increases stably. Especially, we observe that with the same increment of $r_2$, the improvement increases rapidly at the beginning, and then increases slowly. In the case of $r_1=10$, mean IoU  only increases 0.21\% with $r_2$ ranging from 7 to 9. In order to make a trade-off between computational complexity and model accuracy, we keep $0.5<\phi<0.7$ ($\phi$ denotes VFR) in the following experiments.
Secondly, we present the results of varying inter-dilating factor $r_1$ in Tab.~\ref{table:headings1}, where $r_2$ are determined by the principle of $0.5<\phi<0.7$. We also provide the results of atrous convolutions with the same rates for comparison. As the inter-dilating factor $r_1$ increases from 4 to 12, the mean IoU is significantly improved from 76.70\% to 79.79\%. Similar results are observed from atrous convolutions which improve the mean IoU from 75.98\% to 78.18\%. These results show that both the proposed Kronecker convolutions and atrous convolutions benefit from enlarging the field of view, which means Kronecker convolutions can inherit the advantages of atrous convolutions.
Thirdly, we compare the results of Kronecker convolutions and atrous convolutions with the same dilation rates. As shown in Tab.~\ref{table:headings1}, Kronecker convolutions bring 0.8\%, 1.7\%, 0.7\%, 1.5\%, 1.6\% improvements respectively with the dilation rates ranging from 4 to 12. These results show that Kronecker convolutions are stably superior to atrous convolutions, since Kronecker convolutions can aggregate partial detail information neglected by atrous convolutions.

\begin{table}
\centering
\caption{Evaluation results of TFA module on PASCAL VOC 2012 validation set. \textbf{KConv:} employing Kronecker convolution on baseline model 'res4' and 'res5'.}
\label{table:headings2}

\begin{tabular}{lcc}
\toprule[1.5pt]
Method  & \bf{mIoU (\%)} & \bf{Acc (\%)}\\
\midrule
Baseline (Baseline)  &  75.98 & 94.80\\
Baseline + TFA\_S &80.18  &95.56 \\
Baseline + TFA\_L &81.26  &95.83 \\
Baseline + KConv &76.70  & 94.98\\
Baseline + KConv + TFA\_S  & 81.34  &95.96 \\
Baseline + KConv + TFA\_L  &82.85&  96.26  \\

\bottomrule[1.5pt]

\end{tabular}
\end{table}

\begin{figure}[t]
\centering
\includegraphics[width=\linewidth]{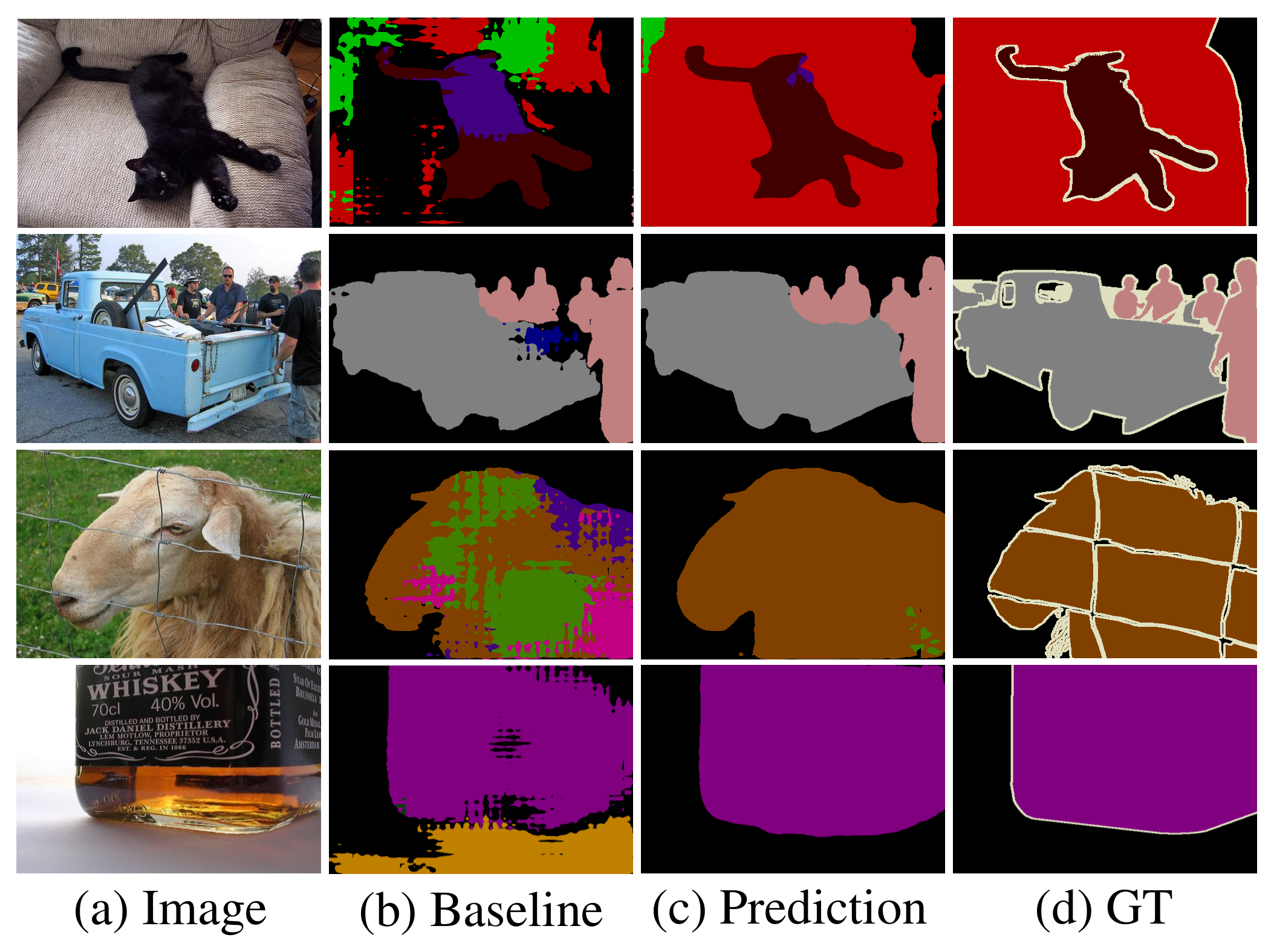}
\vspace{-10pt}
\caption{Result illustration of the proposed TKCN on PASCAL VOC 2012 validation set. From left to right: Input image,  baseline, prediction and ground-truth (GT).}
\vspace{-1pt}
\label{fig:fig5}
\end{figure}

\begin{table}[h]
\rowcolors{1}{blue!6}{blue!0}
\begin{center}
\setlength{\tabcolsep}{1pt}
\caption{Per-class mean intersection-over-union (IoU) results on the PASCAL VOC 2012 segmentation challenge test set, only using VOC 2012 for training. \textbf{Ms:} employing multi-scale inputs with average fusion during testing.}

\label{table:headings3}
\begin{tabular}{p{3cm}p{3cm}<{\centering}}
\toprule[1.5pt]
 \bf {Method} & \bf{mIoU (\%)}\\
\midrule
FCN \cite{shelhamer2017fully}& 62.2\\


GCRF \cite{vemulapalli2016gaussian}   &73.2 \\

Piecewise \cite{lin2016efficient}  & 75.3\\

DeepLab \cite{chen2016deeplab}  & 79.7\\
LC \cite{li2017not}  & 80.3 \\
RAN-s \cite{huang2017semantic}  & 80.5\\
RefineNet \cite{lin2017refinenet}  &82.4\\
{PSPNet\_Ms} \cite{Zhao_2017_CVPR}  &82.6\\
DFN\_Ms \cite{Yu_2018_CVPR}   &82.7\\
EncNet\_Ms \cite{zhang2018context} & 82.9 \\
\midrule
Deeplabv3+ \cite{chen2018encoder} & 89.0 \\
\midrule
{TKCN}  & {82.4} \\
{TKCN\_Ms}  & {83.2}\\
\bottomrule[1.5pt]
\end{tabular}
\end{center}
\vspace{-15pt}
\end{table}

\subsubsection{Ablation Study for TFA Module }

We perform experiments to evaluate our proposed TFA module, which employs Kronecker convolutions in all convolutional layers. We adjust the factors of Kronecker convolutions in the three convolutional layers in TFA module and compare three different schemes: (1) Baseline model of dilated ResNet-101; (2) TFA\_S configured with small factors $(r_1, r_2)=\{(6,3), (10,7), (20,15)\}$ and (3) TFA\_L configured with large factors $(r_1, r_2)=\{(10,7), (20,15), (30,25)\}$. As shown in Tab.~\ref{table:headings2}, compared with baseline, TFA\_S acquires 4.20\% improvement over baseline, while TFA\_L with larger factors bring more improvement of 5.28\%. These results show the effectiveness of TFA module, since hierarchical information can be efficiently aggregated through its tree-shaped structure. Moreover, we implement Kronecker convolutions in the `Res4' and `Res5' layers in the baseline model, denoted as `KConv'. As shown in Tab.~\ref{table:headings2}, KConv+TFA\_S yields $5.36\%$ improvement over baseline and 1.06\% improvement over Baseline + TFA\_S, while KConv+TFA\_L yields $6.87\%$ improvement over baseline and 1.59\% improvement over Baseline + TFA\_L. Therefore, Kronecker convolutions and TFA module can be utilized together to improve the segmentation accuracy cooperatively. In addition, the proposed TFA module has strong generalization capability, since TFA module can bring obvious improvements over both KConv and Baseline.

\begin{figure}
\centering
\includegraphics[width=\linewidth]{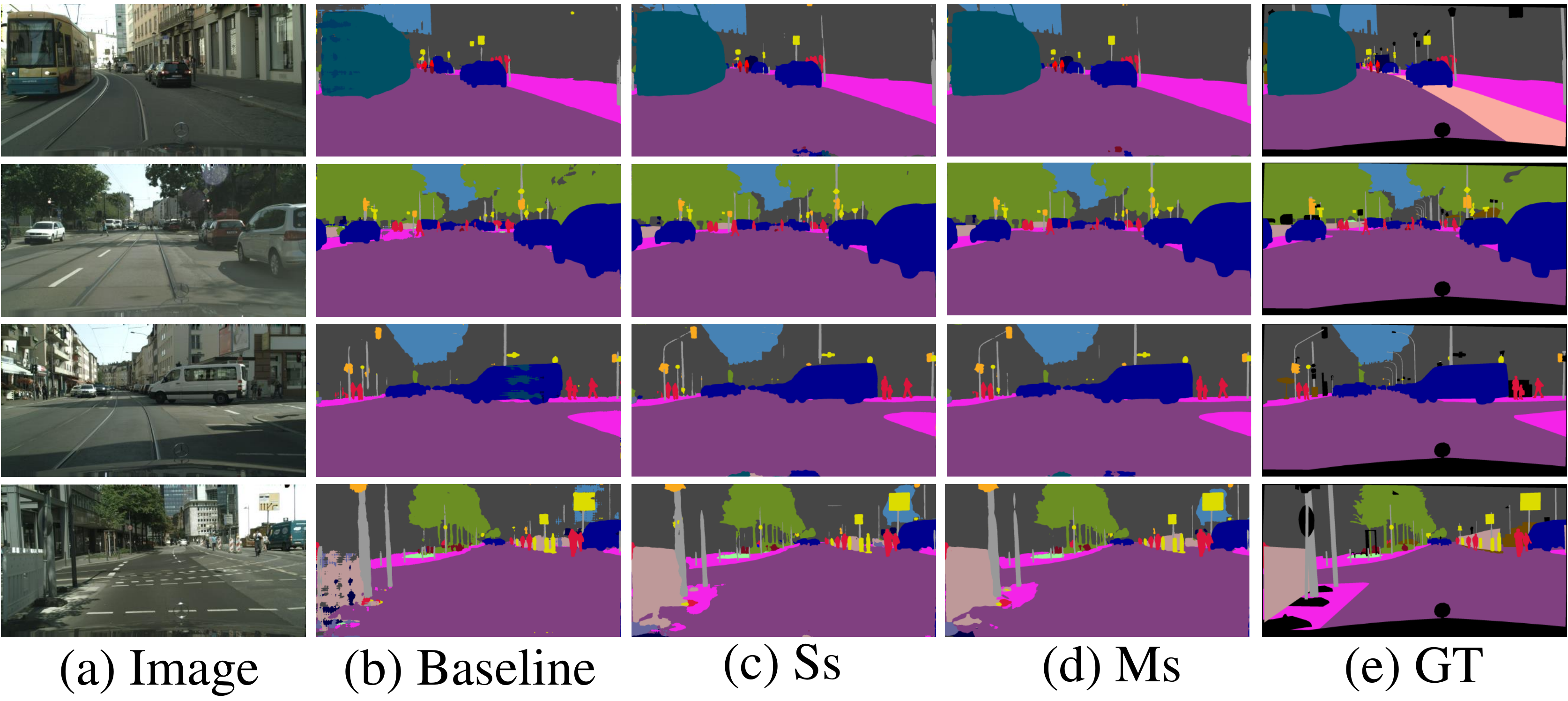}
\caption{Result illustration of the proposed TKCN on Cityscapes validation set. From left to right: Input image, baseline, prediction with single-scale input (Ss), prediction with multi-scale (Ms) input and ground-truth (GT).}
\label{fig:fig7}
\vspace{-10pt}
\end{figure}


\begin{table}[h]
\rowcolors{1}{blue!8}{blue!0}
\begin{center}
\setlength{\tabcolsep}{1.2pt}

\caption{Per-class mean intersection-over-union (IoU) accuracy on Cityscapes test set, only training with the fine set. \textbf{Ms:} employing multi-scale inputs with average fusion during testing.}
\label{table:headings4}
\begin{tabular}{p{3cm}p{3cm}<{\centering}}

\toprule[1.5pt]

\bf{Method} &   \bf{mIoU (\%)} \\
\midrule

CGNet \cite{wu2018cgnet}  & 64.8\\
FCN \cite{shelhamer2017fully}  & 65.3\\





DeepLab \cite{chen2016deeplab} &70.4\\

LC \cite{li2017not}  &71.1\\


RefineNet \cite{lin2017refinenet}  &73.6\\
FoveaNet \cite{Li_2017_ICCV}  &74.1\\
GRLRNet \cite{zhang2017global} &77.3 \\
SAC\_Ms \cite{zhang2017scale} &78.1\\
PSPNet\_Ms \cite{Zhao_2017_CVPR}  & 78.4\\

BiSENet\_Ms \cite{yu2018bisenet} & 78.9\\
DFN\_Ms \cite{Yu_2018_CVPR}   &79.3\\
\midrule
DenseASPP\_Ms \cite{yang2018denseaspp} & 80.6\\
\midrule
TKCN  &{78.9}\\
{TKCN\_Ms} &{79.5}\\
\bottomrule[1.5pt]
\end{tabular}
\end{center}
\vspace{-15pt}
\end{table}

\begin{figure}[h]
\centering
\includegraphics[width=\linewidth]{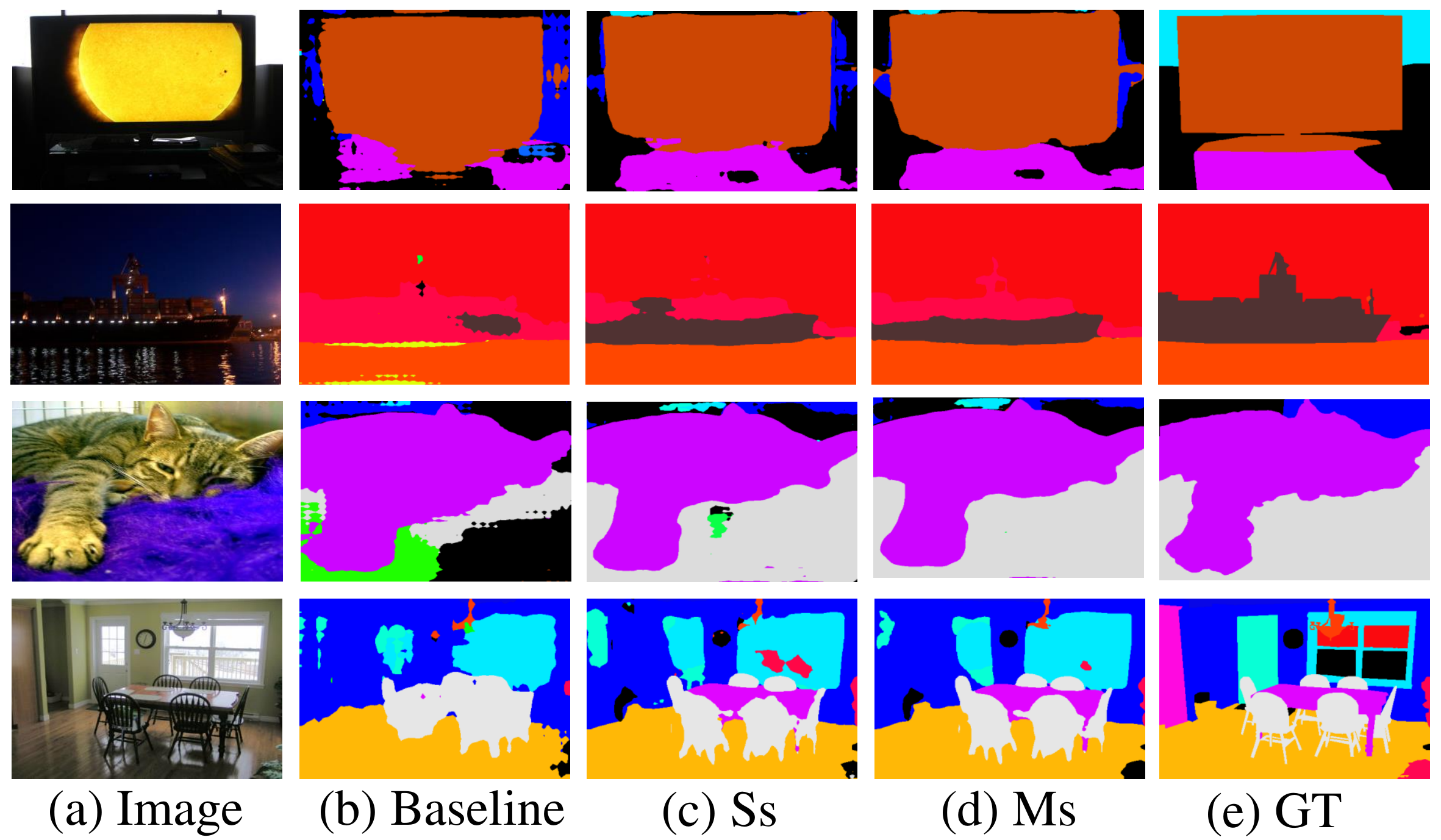}
\vspace{-6pt}
\caption{Result illustration of the proposed TKCN on PASCAL-Context validation set. From left to right: Input image, baseline, prediction with single-scale input (Ss), prediction with multi-scale (Ms) input and ground-truth (GT).}
\label{fig:fig6}
\vspace{-10pt}
\end{figure}

\subsection{Comparison with State-of-the-Arts}
In the following, we present the results of TKCN and compare with other state-of-the-art approaches.

\subsubsection{Results on PASCAL VOC 2012:} We evaluate our proposed TKCN model on PASCAL VOC 2012 dataset without external data such as COCO dataset \cite{caesar2016coco}. Tab.~\ref{table:headings3} shows the results of TKCN compared with other state-of-the-art methods on the test set. Our TKCN method achieves 83.2\% mean IoU (without pre-trained on extra datasets). Our approach is only lower than the famous DeepLabv3+ \cite{chen2018encoder}, which employs a more powerful network (Xception \cite{chollet2017xception}) as the backbone and is pretrained on COCO \cite{caesar2016coco} and JFT \cite{sun2017revisiting}, resulting in 6\% mean IoU improvement.
Fig.~\ref{fig:fig5} displays some qualitative results of the TKCN on the validation set of PASCAL VOC 2012, which shows that the proposed TKCN carries out more accurate and finer structures compared with the baseline.

\begin{table}[h]
\rowcolors{1}{blue!6}{blue!0}
\begin{center}
\setlength{\tabcolsep}{4pt}
\caption{Comparison with other state-of-the-art methods on PASCAL-Context dataset, \textbf{Ms:} employing multi-scale inputs with average fusion during testing.}
\label{table:headings5}
\begin{tabular}{p{3cm}p{3cm}<{\centering}}

\toprule[1.5pt]
\bf{Method}&  \bf{mIoU (\%)} \\

\midrule

FCN \cite{shelhamer2017fully} & 35.1  \\


Context \cite{lin2017exploring}   & 43.5 \\

DeepLab \cite{chen2016deeplab} &  45.7\\

RefineNet\_Ms \cite{lin2017refinenet}   & 47.1 \\
PSPNet\_Ms \cite{Zhao_2017_CVPR} & 47.8  \\
WRNet\_Ms \cite{wu2016wider} & 48.1\\
CCL\_Ms \cite{Ding_2018_CVPR} & 51.6 \\
EncNet\_Ms \cite{zhang2018context} & 51.7 \\
\midrule

{TKCN}  &  {51.1}\\

{TKCN\_Ms} &  {51.8} \\

\bottomrule[1.5pt]
\end{tabular}
\end{center}

\vspace{-15pt}
\end{table}
\subsubsection{Results on Cityscapes:} We report the evaluation results of the proposed TKCN on Cityscapes test set and compare to other state-of-the-art methods in Tab.~\ref{table:headings4},  which shows similar conclusion with the results on PASCAL VOC 2012 dataset. The proposed TKCN achieves 79.5\% in mean IoU (only training on fine annotated images), which is slightly lower than the very recent DenseASPP \cite{yang2018denseaspp} which employs a more powerful network (DenseNet \cite{huang2017densely}) as the backbone.
We visualize some segmentation results on the validation set of Cityscapes in Fig.~\ref{fig:fig7}.

\subsubsection{Results on PASCAL-Context:} Tab.~\ref{table:headings5} reports the evaluation results of proposed TKCN on PASCAL-Context validation set. 
Our proposed model yields $51.1\%$ in mean IoU. Similar to \cite{chen2016deeplab},  employing multi-scale inputs with average fusion further improves the performance to $51.8\%$, which outperforms current state-of-the-art performance. We visualize the prediction results of our model in Fig.~\ref{fig:fig6}.

\section{Conclusions}
In this paper, we propose a novel Kronecker convolution for capturing partial information when enlarging the receptive field of filters. Furthermore, based on Kronecker convolutions, we propose Tree-structured Feature Aggregation module which can effectively capture hierarchical spatial dependencies and learn representations of multi-scale objects. Ablation studies show the effectiveness of each proposed components. Finally, our designed Tree-structured Kronecker Convolutional Network achieves state-of-the-art on the PASCAL VOC 2012, PASCAL-Context and Cityscapes semantic segmentation benchmarks, which demonstrates that our approaches are effective and efficient for high-quality segmentation results.

{\small
\bibliographystyle{unsrt}
\bibliography{egbib}
}

\end{document}